\pdfoutput=1

\documentclass[11pt]{article}

\usepackage{acl}

\usepackage{times}
\usepackage{latexsym}

\usepackage[T1]{fontenc}
\usepackage[utf8]{inputenc}

\usepackage{microtype}

\usepackage{graphicx}
\usepackage{enumitem}
\usepackage{caption, subcaption}
\usepackage{fancyvrb}
\usepackage{amsmath}
\usepackage{amssymb}
\usepackage{multirow}
\usepackage{booktabs}
\usepackage{algorithm}
\usepackage{algpseudocode}  
\usepackage{color}

\newcommand{\red}[1]{\textcolor{red}{#1}}
\newcommand{\blue}[1]{\textcolor{blue}{#1}}

\title{Unified Structure Generation for Universal Information Extraction}

\author{
  Yaojie Lu${}^{1,4,}$\thanks{~ Part of this work was done when Yaojie Lu and Qing Liu interned at Baidu.},
  Qing Liu${}^{1,4,}$\footnotemark[1],
  Dai Dai${}^{3}$,
  Xinyan Xiao${}^{3}$,
  Hongyu Lin${}^{1,}$\thanks{~ Corresponding authors.},
  \\
  \textbf{Xianpei Han}${}^{1,2,5}$,
  \textbf{Le Sun}${}^{1,2,}$\footnotemark[2],
  \textbf{Hua Wu}${}^{3}$
  \\
  ${}^{1}$Chinese Information Processing Laboratory ~
  ${}^{2}$State Key Laboratory of Computer Science \\
  Institute of Software, Chinese Academy of Sciences, Beijing, China\\
  ${}^{3}$Baidu Inc., Beijing, China \\
  ${}^{4}$University of Chinese Academy of Sciences, Beijing, China \\
  ${}^{5}$Beijing Academy of Artificial Intelligence, Beijing, China \\
 {\tt \{yaojie2017,liuqing2020,hongyu,xianpei,sunle\}@iscas.ac.cn} \\
 {\tt \{daidai,xiaoxinyan,wu\_hua\}@baidu.com }
}

\begin{document}
\maketitle
\begin{abstract}
Information extraction suffers from its varying targets, heterogeneous structures, and demand-specific schemas.
In this paper, we propose a unified text-to-structure generation framework, namely UIE, which can universally model different IE tasks, adaptively generate targeted structures, and collaboratively learn general IE abilities from different knowledge sources. 
Specifically, UIE uniformly encodes different extraction structures via a structured extraction language, adaptively generates target extractions via a schema-based prompt mechanism – structural schema instructor, and captures the common IE abilities via a large-scale pre-trained text-to-structure model. 
Experiments show that UIE achieved the state-of-the-art performance on 4 IE tasks, 13 datasets, and on all supervised, low-resource, and few-shot settings for a wide range of entity, relation, event and sentiment extraction tasks and their unification.
These results verified the effectiveness, universality, and transferability of UIE\footnote{\url{https://universal-ie.github.io}}.
\end{abstract}

\section{Introduction} \label{sec:introduction}

Information extraction (IE) aims to identify and structure user-specified information from unstructured texts~\citep{andersen-etal-1992-automatic,grishman_2019}.
IE tasks are highly diversified due to its varying targets (entity, relation, event, sentiment, etc.), heterogeneous structures (spans, triplets, records, etc.), and demand-specific schemas \citep{grishman-sundheim-1996-message,ace2004-annotation,ji-grishman-2011-knowledge}.

Currently, most IE approaches are \textit{task-specialized}, which leads to dedicated architectures, isolated models, and specialized knowledge sources for different IE task.
These task-specialized solutions greatly hinder the rapid architecture development, effective knowledge sharing, and quick cross-domain  adaptation of IE systems.
First, it is very complicated to develop dedicated architectures for a large amount of IE tasks/settings/scenarios.
Second, learning isolated models severely restricts the knowledge sharing between related tasks and settings. 
Finally, it is costly and time-consuming to construct data sets and knowledge sources specialized for different IE tasks.
Therefore, it will be of great benefit to develop a universal IE architecture that can uniformly model different IE tasks, adaptively predict heterogeneous structures and effectively learn from various resources, which we referred to as \textit{\textbf{Universal IE}}.

\begin{figure}[!tpb] 
  \centering
    \includegraphics[width=0.49\textwidth]{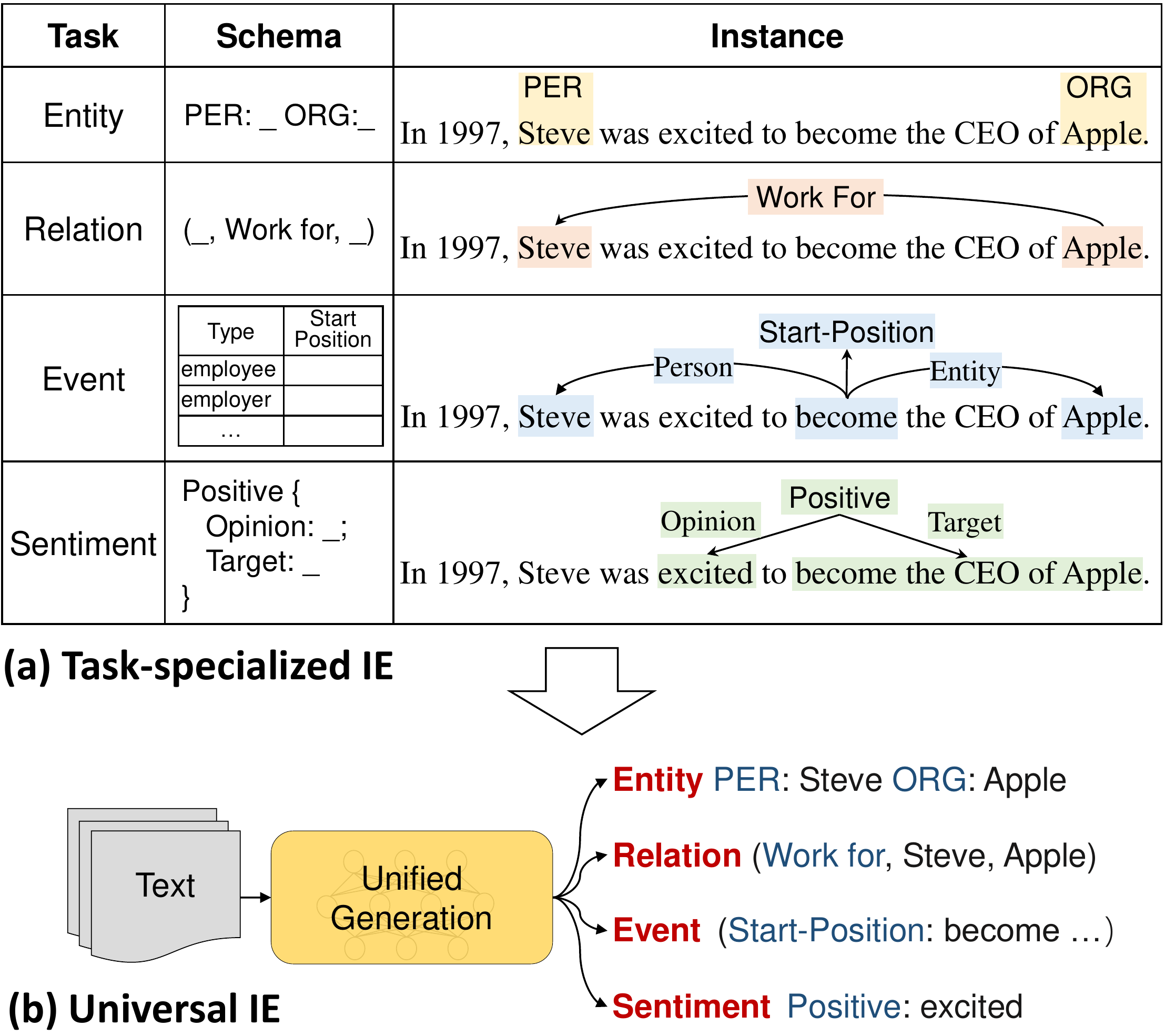}
    \caption{
        From (a) Task-specialized IE: different tasks, different structures, different schemas to (b) Universal IE: unified modeling via structure generation.
    }
    \label{fig:motivation}
\end{figure}%

Fundamentally, all IE tasks can be modeled as text-to-structure transformations, with different tasks correspond to different structures.
For example, as shown in \figurename~\ref{fig:motivation}, an entity is a named span structure, an event is a schema-defined record structure.
These text-to-structure transformations in IE can be further decomposed into several atomic transformation operations:
1) \textit{\textbf{Spotting}}, which locates the desirable spans concerning to given specific semantic types \citep{kripke1971identity,DBLP:conf/cvpr/ChenY04}.
For example, locating span ``Steve'' as a \textit{Person} entity and locating ``excited'' as a sentiment expression.
2) \textit{\textbf{Associating}}, which connects spans by assigning them with semantic roles in pre-defined schemas \citep{onyshkevych-1994-issues,milward-thomas-2000-information}.
For example, associating ``Steve'' and ``Apple'' by assigning them as the  \textit{Arg1} and the \textit{Arg2} of a \textit{Work-for} relation.
In this way, different IE tasks can be decomposed into a sequence of atomic text-to-structure transformations, and all IE models share the same underlying spotting and associating abilities.
For example, entity extraction can be viewed as spotting mention spans of corresponding entity types, while event detection can be reformulated as spotting triggers spans with event types.
And the spotting abilities can be shared between these two tasks.

Based on the above observations, we propose UIE, a unified text-to-structure generation architecture that can universally model different IE tasks, adaptively generate targeted structures, and collaboratively learn general IE abilities from different knowledge sources.
Specifically, to model heterogeneous IE structures, we design a structural extraction language (SEL) that can effectively encode different IE structures into a uniform representation,  so that various IE tasks can be universally modeled in the same text-to-structure generation framework.
To adaptively generate targeted structures for different IE tasks, we propose structural schema instructor (SSI), a schema-based prompt mechanism which controls what to spot, what to associate, and what to generate in UIE.
To learn common IE abilities for UIE, we pre-train UIE on large-scale, heterogeneous datasets mined from easily accessible web sources.
The large-scale pre-trained UIE model provides a solid foundation for knowledge sharing and quick adaptation to new IE settings, and significantly boosts the IE performance in all supervised, low-resource, and few-shot settings.

We conduct experiments on 13 datasets of 4 main IE tasks (entity/relation/event/sentiment extraction and their unification), and supervised, low-resource, and few-shot settings.
Experiment results show that UIE achieves significant improvements in all settings.
On supervised settings, UIE achieved 1.42\% F1 scores improvements over the state-of-the-art, task-specialized architectures on all datasets.
On few-shot and low-resource settings, UIE exhibits strong on-demand adaptation ability: it outperforms baselines dramatically by a large margin.
These results verified the effectiveness,  universality, and transferability of UIE across different IE tasks, settings, and scenarios.

The main contributions of this paper are:

  1) We propose UIE, a unified text-to-structure generation architecture that can universally model different IE tasks, adaptively generate targeted structures, and collaboratively learn general IE abilities from different knowledge sources. 
  
  2) We design a unified structure generation network, which encodes heterogeneous IE structures into a uniform representation via a structural extraction language, and controls the UIE model which to spot, which to associate, and which to generate via structural schema instructor mechanism.
  
  3) We pre-train a large-scale text-to-structure generation model via a unified pre-training algorithm. 
  To the best of our knowledge, this is the first text-to-structure pre-trained extraction model, which can benefit future IE studies.

\section{Unified Structure Generation for Universal Information Extraction} \label{sec:unified}

Information extraction tasks can be formulated as text-to-structure problems, where different IE tasks correspond to different structures.
This paper aims to uniformly model the text-to-structure transformations of different IE tasks via a single framework, i.e., different structure transformations will share the same underlying operations and different transformation abilities in a universal model.
Formally, given a specific pre-defined schema $s$ and texts $x$, a universal IE model needs to generate a structure that contains the desirable structural information in the text $x$ indicated by the schema $s$.

Generally, there are two main challenges here.
Firstly, due to the diversity of IE tasks, there are many different target structures to extract, e.g., entity, relation, event, etc.
Secondly, IE tasks are often demand-specific which are defined using different schemas, therefore we need to adaptively control the extraction process.

In this section, we describe how to jointly formulate, learn, and conduct various IE tasks in a unified text-to-structure generation architecture, named \textbf{UIE}. 
Specifically, we first design structured extraction language (SEL) to uniformly encode heterogeneous extraction structures, i.e., encode entity, relation, event into a unified representation.
Then we  describe structural schema instructor (SSI), a schema-based prompt mechanism that controls the UIE model which to spot, which to associate, and which to generate for different extraction settings. 
The details are as follows.

\subsection{Structured Extraction Language for Uniform Structure Encoding} \label{sec:structured_language}
This section describes how to encode heterogeneous IE structures into a uniform representation.
Based on the above discussions, IE structure generation can be decomposed into two atomic operations:
\begin{enumerate}[nosep,leftmargin=*]
    \item
    \textbf{Spotting} indicates locating target information pieces from the sentence, e.g., the entity and the trigger word in the event.
    \item
    \textbf{Associating} indicates connecting different information pieces based on the desirable associations, e.g., the relation between entity pair or the role between event and its argument.
\end{enumerate}
Then different IE structures can be represented as a combination of atomic structure generation operations.

\begin{figure}[!tpb] 
  \centering

    \begin{subfigure}[b]{0.48\textwidth}
    \begin{minipage}{.48\textwidth}
      \centering
        \begin{Verbatim}[fontsize=\small]
        (
          (Spot Name: Info Span
            (Asso Name: Info Span)
            (Asso Name: Info Span)
          )
         )
        \end{Verbatim}
        \end{minipage}
        \caption{
            Structured extraction language (SEL) for Universal IE.
        }
        \label{fig:sel-illustration}
    \end{subfigure}
    
    \vfill
    
    \begin{subfigure}[b]{0.48\textwidth}
    \begin{minipage}{.96\textwidth}
      \centering
        \begin{Verbatim}[fontsize=\small,commandchars=\\\{\}]
        (
          \blue{(person: Steve}
            \blue{(work for: Apple)}
          \blue{)}
          \red{(start-position: became}
            \red{(employee: Steve)}
            \red{(employer: Apple)}
            \red{(time: 1997)}
          \red{)}
          (organization: Apple)
          (time: 1997)
        )
        \end{Verbatim}
        \end{minipage}
        \caption{
        The SEL representation of the extraction structure of ``Steve became CEO of Apple in 1997.'', where the relation structure is marked \blue{blue}, the event structure is marked \red{red}, and the rest are entities.
        }
        \label{fig:sel-example}
    \end{subfigure}
    
  \setlength{\belowcaptionskip}{-0.4cm}
\caption{
  Illustrations of structured extraction language.
}

\label{fig:sel}
\end{figure}

\begin{figure*}[!tpb]
    \centering
    \includegraphics[width=0.99\textwidth]{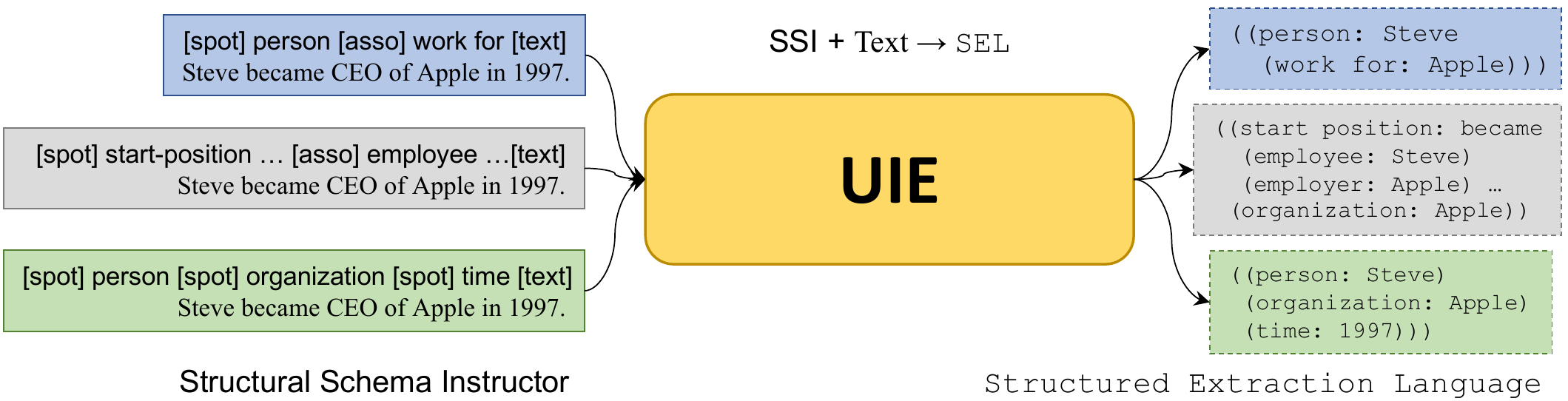}
    \caption{
        The overall framework of UIE.
        }
    \label{fig:full_model}
\end{figure*}

Concretely, we design a unified structured extraction language (SEL), which encodes different IE structures via the spotting-associating structure.
As shown in \figurename~\ref{fig:sel-illustration}, each SEL expression contains three types of semantic units:
1) \textsc{SpotName} represents there is a specific information piece with the type of spot name existing in the source text;
2) \textsc{AssoName} indicates there exists a specific information piece in the source text that is with the AssoName association to its upper-level Spotted information in the structure;
3)  \textsc{InfoSpan} represents the text span corresponding to the specific spotting or associating information piece in the source text.
Furthermore, ``:'' in the SEL indicates the mapping from InfoSpan to its spotting or associating names, and the two structure indicators ``('' and ``)'' are used to form the hierarchical structure between the extracted information.

Using SEL, Figure~\ref{fig:sel-example}  shows how to represent entity, relation, and event structures.
There are three entities and each entity is represented as a spotting structure such as ``person:Steve'', ``organization:Apple'', and ``time:1997'';
one relation which is represented as an association structure between ``Steve'' and ``Apple'' with association name work for;
and one event which is represented as an association structure, where the trigger is a spotting structure ``start-position:became'', and its arguments are associated with the trigger: Steve as employee, Apple as employer, 1997 as time.

We can see that, SEL have the advantages that:
1) uniformly encodes varying IE structures, therefore different IE tasks can be modeled as the same text-to-structure generation process;
2) efficiently represents all extraction results of a sentence in the same structure, thus can perform joint extraction naturally;
3) the output structure of generation is very compact, which greatly reduce the complexity of decoding.

For example, the two different tasks entity recognition and event detection can be revisited using the same ``(SpotName: InfoSpan)'' grammar. 
While both relation extraction and event extraction can be formulated using the grammar ``(SpotName: InfoSpan (AssoName: InfoSpan), ...)'', even they are with totally different binary ``entity-relation-entity'' and N-ary ``event-arguments'' structures.
Such a unified structured extraction language enables UIE to learn from and adapt to different IE tasks without designing task-specialized architectures, because these IE tasks are all universally formulated as the transformation from texts to SEL representations.

\subsection{Structural Schema Instructor for Controllable IE Structure Generation} \label{sec:schema_Instructor}
Using SEL, UIE can uniformly generate different IE structures.
However, because different IE tasks have different schemas, one challenge here is how to adaptively control which information we want to generate during extraction.
For example, given a sentence ``Steve became CEO of Apple in 1997.'', an entity recognition system will generate ``((person: Steve) (organization: Apple) (Time: 1997))'', and an event extraction system will generate ``(start position: became (employee: Steve) (employer: Apple))''.
To this end, we propose structural schema instructor (SSI), a schema-based prompt mechanism that controls which kinds of information need to be spotted and associated.

Figure~\ref{fig:full_model} shows the overall framework of UIE.
Formally, UIE takes the given structural schema instructor ($s$) and the text sequence ($x$) as input, and generates the linearized SEL ($y$) which contains the extracted information from $x$ based on schema $s$:
\begin{equation} \label{equ:uie}
    y = \text{UIE}(s \oplus x)
\end{equation}
where $x=[x_{1}, ..., x_{|x|}]$ is the text sequence, $s=[s_{1}, ..., s_{|s|}]$ is the structural schema instructor, and $y=[y_{1}, ..., y_{|y|}] $ is a SEL sequence that can be easily converted into the extracted information record.

\subsubsection{Structural Schema Instructor}
To describe the extraction target of a task, the structural schema instructor constructs a schema-based prompt and uses it as a prefix during generation.

Specifically, corresponding to the spotting-association structure, the structural schema instructor contains three types of token segments:
1) \textsc{SpotName}: the targeted spotting name in the specific information extraction task, such as ``person`` in the NER task; 
2) \textsc{AssoName}: the targeted association name, such as ``work for'' in the relation extraction task;
3) Special Symbols ([spot], [asso], [text]) which are added before each \textsc{SpotName}, \textsc{AssoName}, and input text sequence.
All tokens in SSI are concatenated and put before the original text sequences.
As shown in Figure~\ref{fig:full_model}, the entire input for UIE is in the form of:
\begin{equation} \label{equ:ssi}
    \begin{aligned}
        s \oplus x & = [s_{1}, s_{2}, ..., s_{|s|}, x_{1}, x_{2}, ..., x_{|x|}] \\
         & = [\text{[spot]}, ...\text{[spot]} ...,\\
         & ~~~~~~~ \text{[asso]}, ..., \text{[asso]} ..., \\
         & ~~~~~~~ \text{[text]}, x_{1}, x_{2}, ..., x_{|x|}]
    \end{aligned}
\end{equation}
For example, the SSI ``[spot] person [spot] company [asso] work for [text]'' indicates extracting records of the relation schema ``the person works for the company'' from the sentence.
Given the SSI $s$, UIE first encodes the text $x$, then generates the target record $y$ in linearized SEL using an encoder-decoder-style architecture.

We found that the schema-based prompt can:
1) effectively guide the SEL generation of UIE, so that the general IE ability can be transferred to new IE tasks;
2) adaptively control which to spot, which to associate, and which to generate, so that semantic knowledge across different labels and tasks can be better shared.

\subsubsection{Structure Generation with UIE}
Given SSI $s$ and text $x$ as input, UIE extracts targeted information by generating a linearized SEL.
We formulate this text-to-SEL generation process using an encoder-decoder-style architecture.
Given the raw text sequence $x$ and the schema instructor $s$, UIE first compute the hidden representation $\mathbf{H} = [\mathbf{s}_{1}, ..., \mathbf{s}_{|s|}, \mathbf{x}_{1}, ..., \mathbf{x}_{|x|}]$ of each token:
\begin{equation} \label{equ:encoder}
    \mathbf{H} = \text{Encoder}(s_{1}, ..., s_{|s|}, x_{1}, ..., x_{|x|})
\end{equation}
where $\text{Encoder}(\cdot)$ is a Transformer encoder.
Then UIE will decode the input text into a linearized SEL in an auto-regressive style.
At the step $i$ of decoding, UIE generates the $i$-th token $y_{i}$ in the SEL sequence and the decoder state $\mathbf{h}_{i}^{d}$ as following:
\begin{equation} \label{equ:decoder}
    y_{i}, \mathbf{h}_{i}^{d} = \text{Decoder}([\mathbf{H}; \mathbf{h}_{1}^{d}, ..., \mathbf{h}_{i-1}^{d}])
\end{equation}
$\text{Decoder}(\cdot)$ is a transformer decoder, which predicts the conditional probability $p(y_{i} | y_{<i}, x, s)$ of token $y_{i}$.
Finally, $\text{Decoder}(\cdot)$ finishes prediction when outputting the end symbol <eos>, then we convert the predicted SEL expression into the extracted information record.

Compared with previous IE studies which treat labels as specific symbols, the text-to-structure generation paradigm treats labels as natural language tokens.
By verbalizing and generating  labels and structures, our method can effectively transfer knowledge from pre-trained language models such as BART \citep{lewis-etal-2020-bart}, T5 \citep{2020t5}, and related tasks can easily share knowledge because their labels have similar semantics (e.g., \textit{location} and \textit{place}) and share common label-text associations (e.g., \textit{victim} for different event types).

\section{Pre-training and Fine-tuning for UIE} \label{sec:learning}

In this section, we describe:
1) how to pre-train a large-scale UIE model which captures common IE abilities for different IE tasks;
2) how to adapt UIE to different IE tasks in different settings via quick fine-tuning.
Specifically, we first collect several large-scale datasets from the Web, including structured (e.g., knowledge bases), unstructured (e.g., raw texts), and parallel (e.g., Wikipedia-Wikidata links) data, then we uniformly pre-train our UIE model on these heterogeneous datasets.
Finally, we adapt the pre-trained UIE model to the specific downstream IE tasks via on-demand fine-tuning.
We found that the pre-trained UIE model provides a solid foundation for capturing, sharing, and transferring knowledge between different IE tasks, and new IE tasks can be effectively solved because UIE learns general IE ability. 

\subsection{Pre-training Corpus Construction}
UIE needs to encode the text, map text to structure, and decode valid structure. 
Therefore, we collect a large-scale pre-training corpus from easily accessible web data sources (more details are in the appendix):

$\mathcal{D}_{\text{pair}}$ is the text-structure parallel data, where each instance is a parallel pair (token sequence $x$, structured record $y$).
We collect large-scale parallel text-structure pairs by aligning Wikidata with English Wikipedia.
$\mathcal{D}_{\text{pair}}$ is used to pre-train the text-to-structure transformation ability of UIE.

$\mathcal{D}_{\text{record}}$ is the structure dataset where each instance is structured record $y$.
We collect structured records from ConceptNet \cite{Speer_Chin_Havasi_2017} and Wikidata.
$\mathcal{D}_{\text{record}}$ is used to pre-train the structure decoding ability of UIE.

$\mathcal{D}_{\text{text}}$ is the unstructured text dataset, and we use all plain texts in English Wikipedia.
$\mathcal{D}_{\text{text}}$ is used to pre-train the semantic encoding ability of UIE.

\subsection{Pre-training}
\label{sec:tasks_for_pretraining}

We pre-train UIE using three sequence generation tasks with above mentioned pre-training datasets.

\paragraph{Text-to-Structure Pre-training using $\mathcal{D}_{\text{pair}}$.}
To capture the fundamental text-to-structure mapping ability, we pre-train UIE using $\mathcal{D}_{\text{pair}} = \{(x ,y)\}$.
Specifically, for each parallel pair ($x$, $y$), we extract the spot type $s_{s+}$ and the associating type $s_{a+}$ in the record $y$ as the positive schema $s_{+} = s_{\text{s+}} \cup s_{\text{a+}}$.
However, we found that if we only feed UIE with a positive schema, it will only simply remember the triplet in the pre-training data.
To learn general mapping ability, we also automatically construct negative schemas for each pair, i.e., we first sample negative spots $s_{\text{s-}}$ and negative association set $s_{\text{a-}}$, then concatenate meta-schema $s_{\text{meta}} = s_{+} \cup s_{\text{s-}} \cup s_{\text{a-}}$, and construct the final extraction target.
For example, \textit{person} and \textit{work for} is the positive schema in the record ``((person: Steve (work for: Apple)))'', and we sample \textit{vehicle} and \textit{located in} as the negative schema to construct meta-schema.
Finally, the objective of  text-to-structure pre-training is:
\begin{equation}  \label{equ:meta_translation_modeling}
  \mathcal{L}_{\text{Pair}} = \sum_{(x,y)\in\mathcal{D}_{\text{pair}}}
  - \log p(y|x,s_{meta};\mathbf{\theta}_{e},\mathbf{\theta}_{d})
\end{equation}
where $\theta_{e}$ and $\theta_{d}$ are the parameter of encoder and decoder, respectively.

\paragraph{Structure Generation Pre-training with $\mathcal{D}_{\text{record}}$.}

To pre-train the ability of generating valid structures defined by SEL and schemas, we pre-train UIE on $\mathcal{D}_{\text{record}}$.
We pre-train UIE decoder as an structured language model, where each record in $\mathcal{D}_{\text{record}}$ is an expression of SEL:
\begin{equation}  \label{equ:structured_language_modeling}
  \mathcal{L}_{\text{Record}} = 
    \sum_{y \in \mathcal{D}_{\text{record}}}
    - \log p(y_{i}| y_{<i}; \mathbf{\theta}_{d})
\end{equation}
By pre-training for structure generation, the decoder can capture the regularity of SEL and the interactions between different labels.

\paragraph{Retrofitting Semantic Representation using $\mathcal{D}_{\text{text}}$.}

During text-to-structure pre-training, we continually pre-train UIE also with the masked language model tasks \citep{2020t5} on $\mathcal{D}_{\text{text}}$ to retrofit semantic representations of UIE.
Specifically, we add span corruption based mask language modeling objective in the pre-training stage:
\begin{equation} \label{equ:mask_language_modeling}
    \mathcal{L}_{\text{Text}} =  \sum_{x \in \mathcal{D}_{\text{text}}} - \log p(x''|x'; \mathbf{\theta}_{e}, \mathbf{\theta}_{d})
\end{equation}
where $x'$ is the corrupted source text and $x''$ is corrupted target spans.
We found this pre-training can effectively alleviate the catastrophic forgetting of token semantics especially on \textsc{SpotName} and \textsc{AssoName} tokens.

\paragraph{Final Pre-training Criteria.}
We initialize UIE-base and UIE-large with T5-v1.1-base and T5-v1.1-large \citep{2020t5}, and the model architectures are shown in \tablename~\ref{tab:model-architectures}.
The final objective is the combine of the above tasks:
\begin{equation}
    \mathcal{L} = \mathcal{L}_{\text{Pair}} + \mathcal{L}_{\text{Record}} + \mathcal{L}_{\text{Text}}
\end{equation}
For implementation, we uniformly represent all pre-training data as triplets.
For text data ($x$) in $\mathcal{D}_{\text{text}}$, we build a triplet (None, $x'$, $x''$) where  $x'$ is the corrupted source text and $x''$ is corrupted spans.
For text-record data ($x$, $y$) in $\mathcal{D}_{\text{pair}}$, we construct ($s$, $x$, $y$) by sampling the meta-schema $s$ for each text-record pair.
For record data ($y$) in $\mathcal{D}_{\text{record}}$, we take (None, None, $y$) as the input triplet.
We randomly pack instances for different tasks in one batch, and details are shown in Algorithm~\ref{alg:uie-pseudocode} in the appendix.

\subsection{On-Demand Fine-tuning}
Given the pre-trained UIE model, we can quickly adapt it to different IE tasks and settings through model fine-tuning.
Given a labeled corpus $\mathcal{D}_{\text{task}} = \{(s, x, y)\}$, we fine-tune the UIE model using teacher-forcing cross-entropy loss:
\begin{equation} \label{equ:fine_tuning}
  \mathcal{L}_{\text{FT}} = \sum_{(s,x,y)\in\mathcal{D}_{\text{Task}}}
  - \log p(y|x,s;\mathbf{\theta}_{e},\mathbf{\theta}_{d})
\end{equation}

\begin{table}[t]
  \centering
  \setlength{\belowcaptionskip}{-0.2cm}
  \resizebox{0.48\textwidth}{!}{
      \begin{tabular}{l|l}
        \toprule
        SSI & <spot> person ... <spot> facility <asso> ... <text> \\
        Text & Steve became CEO of Apple in 1997. \\
        \midrule
        SEL & ((person: Steve (work for: Apple)) (start-position: ... \\
        ~ + RM & ((person: Steve (work for: Apple)) \red{\textit{(facility: \textsc{[null]})}} ... \\
        \bottomrule
      \end{tabular}%
  }
  \caption{
  An example of rejection mechanism (RM), here ``(facility: \textsc{[null]})'' is the injected rejection noise during learning stage, and the \textsc{[null]}-valued span will be ignored during inference stage.
  }
  \label{tab:example_of_rejection_noise}%
\end{table}%

To alleviate the exposure bias \citep{RanzatoCAZ15,zhang-etal-2020-minimize} of the auto-regressive model during decoding, we also design a \textit{\textbf{Rejection Mechanism}} for effective fine-tuning.
Specifically, given an instance ($s$, $x$, $y$), we first encode $y$ using SEL language, then we randomly insert several \textsc{[null]} unit with negative \textsc{SpotName} and \textsc{AssoName}: \textsc{(SpotName, [null])} and \textsc{(AssoName, [null])} into the ground-truth SEL with the probability of $p_{\epsilon}$.
For example, in \tablename~\ref{tab:example_of_rejection_noise}, \textsc{$facility$} is the negative spot in the schema prompt, i.e., there is no \textit{facility} entity in the sentence ``Steve became CEO of Apple in 1997''.
Therefore, we randomly inject the noise of ``(facility: \textsc{[null]})'' into the target record during model learning.
In this way, the UIE can effectively learn to reject misleading generation by generating \textsc{[null]} token.

\section{Experiments} \label{sec:experiments}
To verify the effectiveness of UIE, we conducted experiments on different IE tasks and settings.

\subsection{Experimental Settings} \label{sec:uie-benchmark}

\begin{table*}[t]
\centering
\resizebox{0.99\textwidth}{!}{

\begin{tabular}{ccc|cc|cc}
\toprule
\textbf{Dataset} & \textbf{Domain} & \textbf{Metric} & \multicolumn{2}{c|}{\textbf{Comparable SOTA}} & \textbf{SEL} & \textbf{UIE} \\

\midrule
ACE04 & News, Speech & Entity F1 & \citep{yan-etal-2021-unified-generative} & \textbf{86.84} & 86.52  & \textbf{86.89} \\
ACE05-Ent & News, Speech & Entity F1 & \citep{yan-etal-2021-unified-generative} & 84.74  & 85.52  & \textbf{85.78} \\
CoNLL03 & News  & Entity F1 & \citep{wang-etal-2021-improving} & \textbf{93.21} & 92.17  & 92.99  \\

\midrule
ACE05-Rel & News, Speech & Relation Strict F1 & \citep{zhong-chen-2021-frustratingly} & 65.60  & 64.68  & \textbf{66.06} \\
CoNLL04 & News & Relation Strict F1 & \citep{wang-lu-2020-two} & 73.60  & 73.07  & \textbf{75.00} \\
NYT   & News  & Relation Triplet F1 & \citep{zheng-etal-2021-prgc} & 92.70  & \textbf{93.54} & \textbf{-} \\
SciERC & Scientific & Relation Strict F1 & \citep{zhong-chen-2021-frustratingly} & 35.60  & 33.36  & \textbf{36.53} \\

\midrule
\multirow{2}[1]{*}{ACE05-Evt} & \multirow{2}[1]{*}{News, Speech} & Event Trigger F1 & \citep{lin-etal-2020-joint} & 72.80  & 72.63  & \textbf{73.36} \\
      &       & Event Argument F1 & \citep{lin-etal-2020-joint} & \textbf{54.80} & 54.67  & \textbf{54.79} \\
\multirow{2}[1]{*}{CASIE} & \multirow{2}[1]{*}{Cybersecurity} & Event Trigger F1 & \citep{lu-etal-2021-text2event} & 67.51  & 68.98  & \textbf{69.33} \\
      &       & Event Argument F1 & \citep{lu-etal-2021-text2event} & 59.45  & 60.37  & \textbf{61.30} \\
\midrule
14-res & Reviews & Sentiment Triplet F1 & \citep{zhang-etal-2021-towards-generative} & 72.16  & 73.78  & \textbf{74.52} \\
14-lap & Reviews & Sentiment Triplet F1 & \citep{zhang-etal-2021-towards-generative} & 60.78  & 63.15  & \textbf{63.88} \\
15-res & Reviews & Sentiment Triplet F1 & \citep{xu-etal-2021-learning} & 63.27  & 66.10  & \textbf{67.15} \\
16-res & Reviews & Sentiment Triplet F1 & \citep{xu-etal-2021-learning} & 70.26  & 73.87  & \textbf{75.07} \\
\bottomrule
\end{tabular}%

}

\caption{
    Overall results of UIE-large on different datasets. SEL refers to UIE without pre-training by directly using T5-v1.1-large as the backbone.
    Because NYT overlaps with pre-training data, we didn't conduct UIE on NYT for fair comparsion.
    More results of UIE-base and the details of evaluation metric are shown in the appendix.
}
\label{tab:overall}

\end{table*}

\begin{table*}[htbp]
\centering
  \resizebox{0.99\textwidth}{!}{
\begin{tabular}{cl|cccc|cccc}
\toprule
      & \multicolumn{1}{c|}{\textbf{Model}} & \textbf{1-Shot} & \textbf{5-Shot} & \textbf{10-Shot} & \textbf{AVE-S} & \textbf{1\%} & \textbf{5\%} & \textbf{10\%} & \textbf{AVE-R} \\
\midrule
\multicolumn{1}{c}{\multirow{4}[2]{*}{\shortstack{\textbf{Entity} \\ (\textbf{CoNLL03}) \\ \textbf{Ent-F1}}}} & T5-v1.1-base & 12.73  & 30.17  & 58.89  & 33.93  & 75.74  & 85.71  & 87.70  & 83.05  \\
      & Fine-tuned T5-base & 24.93  & 54.85  & 65.31  & 48.36  & 78.51  & 87.67  & 88.91  & 85.03  \\
      & UIE-base w/o SSI & 43.52  & 64.76  & 72.47  & 60.25  & 81.91  & \textbf{88.41} & \textbf{89.84} & 86.72  \\
      & UIE-base & \textbf{46.43} & \textbf{67.09} & \textbf{73.90} & \textbf{62.47} & \textbf{82.84} & 88.34 & 89.63  & \textbf{86.94} \\
\midrule
\multicolumn{1}{c}{\multirow{4}[2]{*}{\shortstack{\textbf{Relation} \\ (\textbf{CoNLL04}) \\ \textbf{Rel-S F1}}}} & T5-v1.1-base & 2.35  & 7.99  & 25.98  & 12.11  & 6.08  & 32.38  & 41.87  & 26.78  \\
      & Fine-tuned T5-base & 4.24  & 28.16  & 41.44  & 24.61  & 12.89  & 37.75  & 49.95  & 33.53  \\
      & UIE-base w/o SSI & 13.21  & 40.35  & 49.47  & 34.34  & 24.21  & 48.70  & 56.59  & 43.17  \\
      & UIE-base & \textbf{22.05} & \textbf{45.41} & \textbf{52.39} & \textbf{39.95} & \textbf{30.77} & \textbf{51.72} & \textbf{59.18} & \textbf{47.22} \\
\midrule
\multicolumn{1}{c}{\multirow{4}[2]{*}{\shortstack{\textbf{Event Trigger} \\ (\textbf{ACE05-Evt}) \\ \textbf{Evt Tri F1}}}} & T5-v1.1-base & 19.40  & 43.35  & 50.57  & 37.77  & 25.59  & 49.47  & 57.18  & 44.08  \\
      & Fine-tuned T5-base & 30.18  & 48.31  & 51.27  & 43.25  & 31.08  & 51.16  & 57.76  & 46.67  \\
      & UIE-base w/o SSI & 32.07  & 48.11  & 51.00  & 43.73  & 32.71  & 53.20  & 59.26  & 48.39  \\
      & UIE-base & \textbf{38.14} & \textbf{51.21} & \textbf{53.23} & \textbf{47.53} & \textbf{41.53} & \textbf{55.70} & \textbf{60.29} & \textbf{52.51} \\
\midrule
\multicolumn{1}{c}{\multirow{4}[2]{*}{\shortstack{\textbf{Event Argument} \\ (\textbf{ACE05-Evt}) \\ \textbf{Evt Arg F1}}}} & T5-v1.1-base & 2.75  & 20.21  & 27.53  & 16.83  & 3.59  & 21.53  & 30.90  & 18.67  \\
      & Fine-tuned T5-base & 6.96  & 25.07  & 30.96  & 21.00  & 7.39  & 24.97  & 33.90  & 22.09  \\
      & UIE-base w/o SSI & 9.31  & 23.99  & 30.31  & 21.20  & 9.57  & 27.25  & 34.18  & 23.67  \\
      & UIE-base & \textbf{11.88} & \textbf{27.44} & \textbf{33.64} & \textbf{24.32} & \textbf{12.80} & \textbf{30.43} & \textbf{36.28} & \textbf{26.50} \\
\midrule
\multicolumn{1}{c}{\multirow{4}[2]{*}{\shortstack{\textbf{Sentiment} \\ (\textbf{16res}) \\ \textbf{Rel-S F1}}}} & T5-v1.1-base & 0.04  & 2.11  & 12.66  & 4.94  & 3.50  & 27.08  & 45.97  & 25.52  \\
      & Fine-tuned T5-base & 6.55  & 21.06  & 29.92  & 19.18  & 18.72  & 39.63  & 51.65  & 36.67  \\
      & UIE-base w/o SSI & 7.79  & 17.77  & 32.07  & 19.21  & 19.14  & 42.76  & 53.44  & 38.45  \\
      & UIE-base & \textbf{10.50} & \textbf{26.24} & \textbf{39.11} & \textbf{25.28} & \textbf{24.24} & \textbf{49.31} & \textbf{57.61} & \textbf{43.72} \\
\bottomrule
\end{tabular}%

}

\caption{
    Low-resource results on end-to-end IE tasks, where \textbf{AVE-S}(hot) and \textbf{AVE-R}(atio) are the averaged performance across 3 few-shot settings and 3 low-resource settings respectively.
}
\label{tab:lowresource}

\end{table*}

\paragraph{Datasets.}
We conduct experiments on 13 IE benchmarks across 4 well-representative IE tasks (including entity extraction, relation extraction, event extraction, structured sentiment extraction) and their combinations (e.g., joint entity-relation extraction).
The used datasets includes ACE04 \citep{ace2004-annotation}, ACE05 \citep{ace2005-annotation}; CoNLL03 \citep{tjongkimsang2003conll}, CoNLL04 \citep{roth-yih-2004-linear}, SciERC \citep{luan-etal-2018-multi}, NYT \citep{10.1007/978-3-642-15939-8_10}, CASIE \citep{Satyapanich_Ferraro_Finin_2020}, SemEval-14 \citep{pontiki-etal-2014-semeval}, SemEval-15 \citep{pontiki-etal-2015-semeval},
SemEval-16 \citep{pontiki-etal-2016-semeval}, see \tablename~\ref{tab:details_datasets} for detail.
We employ the end-to-end setting for all extraction tasks, which takes the raw text as input and directly generates the target structure.

\paragraph{Evaluation.}
We use the same evaluation metrics as all previous methods, and details of metrics are shown in the appendix.
For each fine-tuning experiment, we report the average performance on 3 random seeds.
Because UIE only generates text spans, we map spans to offsets by finding the first matched offsets that are not already matched in the same SEL hierarchical level (details in appendix).
We found this simple heuristic rule is very effective (<0.5\% error offsets) and more complicated mapping approaches (such as attention-weight guided span mapping) are left as the future work.

\subsection{Experiments on Supervised Settings}
UIE provides a universal backbone for IE tasks.
This section assesses the UIE performance in supervised settings.
We compare UIE with the state-of-the-art, task-specific supervised models.
For a fair comparison, we only compare the state-of-the-art without leveraging additional dataset-specific knowledge or larger-scale contexts.
These extensions are good complementary of UIE, and can be left for further improvement.
\tablename~\ref{tab:overall} shows the performance of UIE on the 13 IE datasets across 4 tasks.
We can observe that:

1)
\textit{By modeling IE as text-to-structure generation and encoding with an effective SEL language, UIE provides an effective universal architecture for IE.}
The UIE model achieves state-of-the-art performance on nearly all datasets and tasks, even without pre-training (SEL).
2)
\textit{The large-scale pre-trained model provides a solid foundation for universal IE.}
Compared with baselines, the pre-trained model achieves the performance of the state-of-the-art in most datasets and improves 1.42\% F1 on average.
3) 
\textit{By universally modeling IE tasks and pre-training using large-scale datasets, UIE can effectively capture, share, and transfer IE abilities.}
Pre-training improves all tasks at the same time, especially events and sentiment knowledge rarely appear in the pre-train dataset.
It proves that SEL is a unified and cross-task transferable structured representation for IE, which allows UIE to share learned capabilities and information across different and various information extraction tasks.

\subsection{Experiments on Low-resource Settings}

To verify the quick adaptation ability of UIE, we conducted low-resource experiments on six different partitions of the original training sets (1/5/10-shot, 1/5/10\% ratio) across 4 tasks.
For the few-shot experiments, we sample 1/5/10 sentences for each entity/relation/event/sentiment type in the training set.
To avoid the influence of random sampling, we repeated each experiment 10 times with different samples and reported their averaged results as previous works~\citep{huang-etal-2021-shot}.

We compare UIE with the following pre-trained model: 1) \textbf{T5-v1.1-base} is an initial model of UIE-base; 2) \textbf{Fine-tuned T5-base} is fine-tuned with sequence generation tasks such as summarization, which have been shown effective in many low-resource NLP tasks~\cite{paolini2021structured}; 3) \textbf{UIE-base w/o SSI} is the distant supervised version of UIE without SSI in the pre-training stage, which is used to verify the necessity of SSI when adapting UIE in low-resource settings.
Table~\ref{tab:lowresource} shows the performance of 4 IE tasks under 6 low-resource settings. 
We observe that:
1) \textit{By guiding the generation using schema-based prompts, SSI is an effective way for adaptively controlling which to extract.}
Compared with the UIE model w/o SSI, UIE equipped with SSI achieves improvements of 4.16 and 3.30 on average for n-shot and n-ratio experiments.
2) \textit{Our pre-training algorithms can learn general IE ability rather than capture task-specific information.}
Even the pre-training of UIE didn't include event and sentiment knowledge, UIE still achieved significantly better performance on these tasks compared to the baseline with only a small number of samples.

\subsection{Ablations on Pre-training Tasks}

\begin{table}[htbp]
  \centering
  \resizebox{0.49\textwidth}{!}{
\begin{tabular}{lccccc}
\toprule
\multicolumn{1}{c}{\textbf{Task}} & \textbf{Entity} & \textbf{Relation} & \multicolumn{2}{c}{\textbf{Event}} & \textbf{Sent.} \\
\midrule
\multicolumn{1}{c}{\textbf{F1}} & \textbf{Ent} & \textbf{Rel-S} & \textbf{Evt-Tri} & \textbf{Evt-Arg} & \textbf{Rel-S} \\
\midrule
UIE-base & \textbf{95.89} & \textbf{75.97} & \textbf{72.63} & 57.27  & \textbf{74.73} \\
~ w/o $\mathcal{L}_{\text{Pair}}$ & 95.83  & 75.07  & 71.20  & 55.79  & 74.27  \\
~ w/o $\mathcal{L}_{\text{Record}}$ & 95.69  & 75.68  & 71.99  & \textbf{57.60} & 74.43  \\
~ w/o $\mathcal{L}_{\text{Text}}$ & 95.66 & 75.70  & 70.89  & 54.16  & 74.28  \\
\midrule
T5-v1.1-base & 95.29  & 72.12  & 70.50  & 54.42  & 72.03  \\
\bottomrule
\end{tabular}%

}
  \caption{
  Experiment results of UIE-base with different learning tasks on the development set of four downstream datasets: entity (CoNLL03), relation (CoNLL04), event (ACE05-Evt) and sentiment (16res).
}
  \label{tab:learning_task}%
\end{table}%

\begin{table}[t]
  \centering
    \setlength{\belowcaptionskip}{-0.3cm}
  \resizebox{0.48\textwidth}{!}{
  
    \begin{tabular}{lcccc}
        \toprule
              & $\Delta$ P & P     & R     & F \\
        \midrule
        UIE-base & \multirow{2}[2]{*}{\textbf{+11.41}} & 79.54  & 72.63  & 75.91  \\
        ~ w/o rejection &       & 68.13  & 67.85  & 66.13  \\
        \midrule
        UIE-base w/o SSI & \multirow{2}[2]{*}{\textbf{+9.41}} & 78.96  & 70.50  & 74.49  \\
        ~ w/o rejection &       & 69.55  & 63.69  & 66.44  \\
        \midrule
        T5-base & \multirow{2}[2]{*}{\textbf{+17.95}} & 74.12  & 61.72  & 67.33  \\
        ~ w/o rejection &       & 56.17  & 56.00  & 55.94  \\
        \midrule
        T5-v11 & \multirow{2}[2]{*}{\textbf{+13.88}} & 71.88  & 51.23  & 59.67  \\
        ~ w/o rejection &       & 58.00  & 45.04  & 50.38  \\
        \bottomrule
    \end{tabular}

   }
  \caption{
  Experiment results of 10-shot setting on the CoNLL 03 development set.
}
  \label{tab:rejection_noise}%
\end{table}%

To investigate the effect of different pre-training tasks, \tablename~\ref{tab:learning_task} shows ablation experiment results of UIE-base on four downstream tasks.
We can see that:
(1)
\textit{The pre-training of SEL ($\mathcal{L}_{\text{Record}}$) and sequence-to-structure mapping ($\mathcal{L}_{\text{Pair}}$) is crucial for UIE, and such a structure generation pre-training is especially useful for small-scale datasets.}
In small datasets CoNLL04 and 16res, adding structure generation pre-training (from T5-v1.1-base to UIE-base w/o $\mathcal{L}_{\text{Text}}$), the performance significantly increases from 72.12 to 75.70 and 72.03 to 74.28.
(2)
\textit{Retrofitting semantic using the mask language model task ($\mathcal{L}_{\text{Text}}$) is more important for the complex extraction task.}
In the tasks with more semantic types such as event extraction (33 types), the performance drops significantly after removing the $\mathcal{L}_{\text{Text}}$ task, e.g., 72.63$\rightarrow$70.89 and 57.27$\rightarrow$54.16.
(3) 
\textit{The mapping pre-training with $\mathcal{L}_{\text{Pair}}$ enables the model to learn the ability of extraction.}
After ablating $\mathcal{L}_{\text{Pair}}$, the extraction ability of UIE is significantly decreased, i.e., the performance on the relation (-0.90), event (-1.43/-1.48), and sentiment (-0.46) tasks all see large decline.

\subsection{Effects of Rejection Noise}

This section investigates the effect of the proposed rejection noise.
\tablename~\ref{tab:rejection_noise} shows the results of the different pre-trained models on the development set of CoNLL 03 under the 10-shot setting.
The mis-generated label has a negative influence on the precision of the proposed generation method leading to a large number of error extraction results.
The proposed rejection noise is useful for the generation method, which leads to improvements of 13.16 precision (P) on average.

\section{Related Work} \label{section:related_work}
Building and pre-training universal models of NLP tasks has attracted a lot of attention in recent years, e.g., contextualized representation \citep{devlin-etal-2019-bert, roberta}, text generation \citep{lewis-etal-2020-bart,2020t5}, multi-modal \citep{li-etal-2021-unimo,pmlr-v139-cho21a}, and multi-lingual \citep{conneau-etal-2020-unsupervised,xue-etal-2021-mt5}.
This paper proposes and pre-trains the first universal model for information extraction.

IE is a long-researched area and many classical neural architectures have been proposed, such as sequence tagging \citep{lample-etal-2016-neural,zheng-etal-2017-joint,lin-etal-2019-sequence}, span classification \citep{sohrab-miwa-2018-deep,lin-etal-2018-nugget,wadden-etal-2019-entity}, and MRC \citep{levy-etal-2017-zero,li-etal-2020-unified,du-cardie-2020-event}.
And several task-specific pre-training techniques are proposed on these architectures \citep{mengge-etal-2020-coarse,wang-etal-2021-cleve, qin-etal-2021-erica}.
More relevant to our work are generation-based IE methods, which generate text spans via tagging \citep{strakova-etal-2019-neural,ma-etal-2019-exploring}, index pointer \citep{ren-etal-2021-hyspa,yan-etal-2021-unified-generative} or copy mechanism \citep{zeng-etal-2018-extracting}, and these methods usually employ specific classifiers to represent labels.
The generation can be enhanced using label templates \citep{li-etal-2021-document,liu-etal-2021-fine,cui-etal-2021-template}, schema \citep{lu-etal-2021-text2event,ahmad-etal-2021-intent}, and augmented language methods \citep{paolini2021structured}.

Compared with previous IE studies which focus on developing more effective task-specialized models, this paper aims to universally model various IE tasks in an unified text-to-structure framework, which can greatly benefit the rapid development, effective knowledge sharing, and quick adaptation of IE systems.

\section{Conclusion} \label{sec:conclusion}

In this paper, we propose a unified text-to-structure generation framework – UIE, which can universally model different IE tasks, adaptively generate targeted structures, and unfiedly learn general IE abilities from different knowledge sources.
Experimental results show that UIE achieves very competitive performance in both supervised and low-resource settings, which verified its universality, effectiveness, and transferability.
A large-scale pre-trained text-to-structure model is also released, which will benefit future studies. For future work, we want to extend UIE to KB-aware IE tasks such as entity linking \citep{cao2021autoregressive}, and document-aware IE tasks such as co-reference \citep{lee-etal-2017-end, LU2022103632}.

\section*{Acknowledgements}
We sincerely thank the reviewers for their insightful comments and valuable suggestions.
This research work is supported by the National Natural Science Foundation of China under Grants no. U1936207, 62122077 and 62106251, the Project of the Chinese Language Committee under Grant no. YB2003C002.

\bibliography{custom}
\bibliographystyle{acl_natbib}

\newpage
\appendix

\section{Experiment Details} \label{sec:experiment_data}

This section describes the details of experiments, including pre-training and fine-tuning on downstream tasks.

\subsection{Pre-training Details} \label{sec:pretrain_details}
\paragraph{Data Construction}
We use the 20210401 version of Wikipedia\footnote{\url{https://www.wikipedia.org/}} and Wikidata\footnote{\url{https://www.wikidata.org/}} dump and ConceptNet\footnote{\url{https://conceptnet.io/}} to construct the pre-train dataset.

For Wikidata and Wikipedia, we use them to collect the tuples $\mathcal{T}_w = \{<T_h, e_h, r, e_t, X> \}$, where $T_h$ is head entity type, $e_h$ is head entity, $r$ is relation, $e_t$ is tail entity, $X$ is sentence, and the $\mathcal{T}_w$ can be used to construct $\mathcal{D}_{\text{pair}}$, $\mathcal{D}_{\text{record}}$ and $\mathcal{D}_{\text{text}}$. Firstly, we construct entity type dictionary $\mathcal{L}$ and relation dictionary $\mathcal{P}$ from Wikidata. Wikidata has more than 40M entity items and each item has its corresponding properties which indicate the association between entities. For type dictionary $\mathcal{L}$, we regard each item as an entity, use the “instance of” and “subclass of” property values as its corresponding entity types and consider other properties as the relation of the entity with others. To learn general knowledge, all entity types will be retained except those whose instances are < 5. For the type whose name is longer than 3 tokens, we use its headwords as the final type for simplicity, e.g.,“state award of the Republic of Moldova” is converted to “state award”. For relation dictionary $\mathcal{P}$, Wikidata has more than 9K kinds of properties\footnote{\url{https://www.wikidata.org/wiki/Wikidata:List_of_properties}}, we filter out the properties of external-id, URL, and math types. In this way, we obtain a collection of 31K types and retained 1535 properties which can serve as a solid foundation for universal IE. Secondly, we collect the mentions of each entity by using its anchor texts in Wikipedia and the top 3 frequent noun phrase occurrences of its entry page~\citep{li-etal-2010-generating}. Then for each mention, we identify its entity types by linking it to its Wikidata item's types. For each Wikipedia page, we split the text into sentences\footnote{nltk.tokenize.punkt} and filter out sentences that have no entities. Thirdly, we regard each entity as a head entity and find the associated entities according to its properties. The associated entity will set as as tail entity, and the property value will set as association type. If a head entity has no type, $T_h$ will be blank or has no associated tail entity, $r$ and $e_t$ will be blank. To this end, given a sentence, we can construct instances based on the collected tuples $\mathcal{T}_w$ by setting $e_h$ and $e_t$ as \textsc{InfoSpan}, and assigning $T_h$ as \textsc{SpotName}, $r$ as \textsc{AssoName}. Finally, from Wikipedia and Wikidata, we construct $\mathcal{D}_{\text{pair}}$, $\mathcal{D}_{\text{record}}$ and $\mathcal{D}_{\text{text}}$ with 65M instances, respectively. And we keep 50K as the development dataset. 

To add common sense knowledge to structured extraction language (SEL), we extract the tuples $\mathcal{T}_c$ from ConceptNet. ConceptNet contains 48 associations and has no context or entity types. So we leave the $T_h$, $T_t$ $X$ blank and finally construct 1M instances.

\begin{table*}[htbp]
  \centering
    
  \resizebox{0.98\textwidth}{!}{
  
\begin{tabular}{c|c|cc|c|c|cc}
    \toprule
    \multirow{3}[6]{*}{\textbf{Hyper-parameter}} & \multicolumn{4}{c|}{\textbf{UIE-base}} & \multicolumn{3}{c}{\textbf{UIE-large}} \\
    \cmidrule{2-8}      & \multirow{2}[4]{*}{\textbf{Pre-training}} & \multicolumn{3}{c|}{\textbf{Fine-tuning}} & \multirow{2}[4]{*}{\textbf{Pre-training}} & \multicolumn{2}{c}{\textbf{Fine-tuning}} \\
    \cmidrule{3-5}\cmidrule{7-8}      &       & \textbf{Ent/Rel/Evt} & \multicolumn{1}{c}{\textbf{Sentiment}} & \textbf{Low-resource} &       & \textbf{Ent/Rel/Evt} & \textbf{Sentiment} \\
    \midrule
    Learning Rate & 1e-4  & \multicolumn{2}{c|}{1e-4, 3e-4, 5e-4} & 1e-4  & 1e-4  & \multicolumn{2}{c}{ 5e-5, 1e-4, 3e-4} \\
    Rejection Noise $p_{\epsilon}$ & 0.0 & \multicolumn{2}{c|}{0, 0.1, 0.2} & 0.1   & 0.0 & \multicolumn{2}{c}{0, 0.1, 0.2} \\
    Global Batch Size & 512   & 64    & 16    & 16    & 512   & 32    & 8 \\
    Schedule & linear & linear & linear & constant & linear & linear & linear \\
    Warmup Rate & 0.06  & 0.06  & 0.06  & 0.0  & 0.06  & 0.06  & 0.06 \\
    Epoch/Step & 500K step & 50 epoch & 50 epoch & 200 epoch & 500K step & 50 epoch   & 50 epoch \\
    \bottomrule
\end{tabular}%

  }
  \caption{Hyper-parameters pre-training and fine-tuning for UIE-base and UIE-large.}
  \label{tab:hyper-fine-tuning}%
\end{table*}

\begin{table}[htbp]
  \centering
    \setlength{\belowcaptionskip}{-0.3cm}
  \resizebox{0.45\textwidth}{!}{
    \begin{tabular}{ccc}
    \toprule
    \textbf{Hyper-parameter} & \textbf{UIE-base} & \textbf{UIE-large} \\
    \midrule
    \# Layers of Encoder & 12    & 24 \\
    \# Layers of Decoder & 12    & 24 \\
    Hidden Dimension & 768   & 1,024 \\
    FF hidden size & 2,048 & 2,816 \\
    Layer Normalize $\epsilon$ & 1e-6  & 1e-6 \\
    \# Attention head & 12    & 16 \\
    Attention head size & 64    & 64 \\
    \bottomrule
    \end{tabular}%
  }
  \caption{Model architectures.}
  \label{tab:model-architectures}%
\end{table}%

\newcommand{\pythonstylecomment}[1]{ \leftline{\color{gray}{\# \texttt{\scriptsize{#1}}}}}
\newcommand{\pythonstylefunction}[1]{ \mathbf{#1}}
\begin{algorithm}[t]
	\caption{The pre-training process of UIE in a Python-like style.}  
	\label{alg:uie-pseudocode} 
	\begin{algorithmic}
	\small{
		\State \leftline{\color{gray}{\# \texttt{\scriptsize{The training details of UIE}}}}
		\Function {$\pythonstylefunction{pretraining\_ process}$}{}
		\For{$step$ in $all\_steps$}
		\State $batch = []$

		\State \leftline{\color{gray}{\# \texttt{\scriptsize{load $n_{\text{text}}$ unstructured text samples}}}}
		\State $texts = get\_data(\mathcal{D}_{\text{text}},\ n_{\text{text}})$
		\State \pythonstylecomment{construct corrupted source text $x'$ and}
		\State \pythonstylecomment{corrupted spans $x''$ for each text sample}
		\For{$x$ in $texts$}
		    \State $x', x'' = span\_corrupt(x)$
    		\State $batch.extend((\text{None}, x', x''))$
		\EndFor

		\State \pythonstylecomment{load $n_{\text{record}}$ structured record samples}
		\State $records = get\_data(\mathcal{D}_{\text{record}},\ n_{\text{record}})$
		\For{$y$ in $records$}
    		\State $batch.extend((\text{None}, \text{None}, y))$
		\EndFor
		
		\State \pythonstylecomment{load $n_{\text{pair}}$ text-record parallel pairs}
		\State $text\_record\_pairs = get\_data(\mathcal{D}_{\text{pair}},\ n_{\text{pair}})$
		\State \pythonstylecomment{construct meta-schema $s_{\text{meta}}$}
		\State \pythonstylecomment{for each text-record pair ($x$, $y$)}
		\For{$(x, y)$ in $text\_record\_pairs$}
    		\State$s = \mathbf{meta\_schema\_sample}(y)$
    		\State$batch.extend((s, x, y))$
		\EndFor

		\State \pythonstylecomment{compute loss and backward}
		\State $\mathcal{L}_{\text{Pair}},\ \mathcal{L}_{\text{Record}},\ \mathcal{L}_{\text{Text}} =\ $UIE$(batch)$
		\State $loss = \mathcal{L}_{\text{Pair}} + \ \mathcal{L}_{\text{Record}} + \ \mathcal{L}_{\text{Text}}$
		\State $loss.backward()$
		\EndFor
		\EndFunction
		\\
		\State \pythonstylecomment{The meta sample of UIE}
		\Function {$\pythonstylefunction{meta\_schema\_sample}$}{y}
		\State \pythonstylecomment{get positive spots and associations}
		\State \pythonstylecomment{in the record $y$}
		\State $s_{\text{s+}}, s_{\text{a+}} = get\_schema\_from\_record(y)$
		\State \pythonstylecomment{sample negative spots}
		\State $s_{\text{s-}} = sample\_negative\_spot(s_{+})$
		\State \pythonstylecomment{sample negative associations}
		\State $s_{\text{a-}} = sample\_negative\_association(s_{+})$
		\State $return ~ s_{\text{s+}} \cup s_{\text{s-}} \cup s_{\text{a+}} \cup s_{\text{a-}}$
		\EndFunction
		
	}
	\end{algorithmic}  
\end{algorithm}

\begin{table}[t]
  \centering
  \resizebox{0.49\textwidth}{!}{
    \begin{tabular}{c|ccc|ccc}
    \toprule
          & |Ent| & |Rel| & |Evt| & \#Train & \#Val & \#Test \\
    \midrule
    ACE04 & 7     & -     & -     & 6,202  & 745   & 812  \\
    ACE05-Ent & 7     & -     & -     & 7,299  & 971   & 1,060  \\
    CoNLL03 & 4     & -     & -     & 14,041  & 3,250  & 3,453  \\
    ACE05-Rel & 7     & 6     & -     & 10,051  & 2,420  & 2,050  \\
    CoNLL04 & 4     & 5     & -     & 922   & 231   & 288  \\
    NYT   & 3     & 24    & -     & 56,196  & 5,000  & 5,000  \\
    SciERC & 6     & 7     & -     & 1,861  & 275   & 551  \\
    ACE05-Evt & -     & -     & 33    & 19,216  & 901   & 676  \\
    CASIE & 21     & -     & 5     & 11,189  & 1,778  & 3,208  \\
    14res & 2     & 3     & -     & 1,266  & 310   & 492  \\
    14lap & 2     & 3     & -     & 906   & 219   & 328  \\
    15res & 2     & 3     & -     & 605   & 148   & 322  \\
    16res & 2     & 3     & -     & 857   & 210   & 326  \\
    \bottomrule
    \end{tabular}%
  }
  \caption{
    Detailed datasets statistics.
    |*| indicates the number of categories, and \# is the number of sentences in the specific subset.
    We take sentiment types as special relation type: positive, negative, and neutral; and each sentiment triplet holds a aspect and a opinion.
}
  \label{tab:details_datasets}
\end{table}%

\paragraph{Training Details}

We first initialize UIE-base and UIE-large with T5-v1.1-base and T5-v1.1-large checkpoints~\citep{2020t5}, and the model architectures are shown in \tablename~\ref{tab:model-architectures}.
We employ Adam optimizer \cite{Kingma2015adam} as the optimizer with learning rate=1e-4, and use linear scheduling with a warming up proportion 6\%.
For negative spots and associations in the $\mathcal{L}_{\text{Pair}}$, we randomly select negative spots and associations up to 10 for each instance, respectively.
For $\mathcal{L}_{\text{Text}}$, we set the corruption rate to 15\% and the average corrupting span length to 3, following \citet{2020t5}.
We truncate the concatenated overall length of schema prompt $s$ and raw text $x$, as well as the length of SEL expression $y$, together to 128 during pre-training.
We train our base model and large model for both 500K steps with batch size 512 on 8 NVIDIA A100 GPUs.

The detailed pre-training process in a python-like style is shown in Algorithm \ref{alg:uie-pseudocode}.
In each batch of pre-training processes for UIE, we construct a batch of triplets ($s$, $x$, $y$) containing text-record pairs, text instances, and record instances.
In practice, since 8 GPUs could only run the large model with an overall batch of 128 (batch=16 on each GPU), we update the model parameters after accumulating 4 gradients.

\subsection{Details of Downstream Tasks} \label{sec:downstream_tasks}

We conduct downstream tasks on 4 IE tasks, 13 datasets, and the detailed statistic of each dataset is shown in \tablename~\ref{tab:details_datasets}.

\paragraph{Entity}
We conduct entity extraction experiments on three entity datasets: 
ACE04\footnote{https://catalog.ldc.upenn.edu/LDC2005T09} \citep{ace2004-annotation}, ACE05-Ent\footnote{https://catalog.ldc.upenn.edu/LDC2006T06} \citep{ace2005-annotation}, 
and CoNLL03 \citep{tjongkimsang2003conll}.
For nested entity extraction datasets ACE04 and ACE05-Ent, we follow the pre-processing steps and data split of previous works \cite{li-etal-2020-unified}.

\paragraph{Relation}
We conduct experiments on four wide-used end-to-end relation extraction datasets across several languages and domains: ACE05-Rel \citep{ace2005-annotation}, CoNLL04\footnote{https://github.com/btaille/sincere} \citep{roth-yih-2004-linear}, NYT\footnote{https://github.com/yubowen-ph/JointER} \citep{10.1007/978-3-642-15939-8_10}, and SciERC\footnote{http://nlp.cs.washington.edu/sciIE/} \citep{luan-etal-2018-multi}.
We follow the preprocessing steps and data split of previous works \citep{taille-etal-2020-lets,yu2020jointer,wadden-etal-2019-entity}.

\paragraph{Event}
For ACE05-Evt, we follow the same types, data splits, and pre-processing steps as \citet{lin-etal-2020-joint}.
For CASIE \citep{Satyapanich_Ferraro_Finin_2020}, we first remove three incomplete annotated documents (999, 10001, 10002), then split the remaining documents into three sets: train/val/test=697/100/200 according to the time order of each document.
We employ the state-of-the-art generation-based event extraction method \textsc{Text2Event} \citep{lu-etal-2021-text2event} as the comparable state-of-the-art system.

\paragraph{Sentiment}
We conduct sentiment extraction experiments on the sentiment triplet extraction \citep{xu-etal-2020-position} of SemEval 14/15/16 aspect sentiment analysis datasets.
We employ the pre-processing datasets of the previous work \citep{yan-etal-2021-unified}\footnote{https://github.com/yhcc/BARTABSA}.

\paragraph{Evaluation}
We use span-based offset Micro-F1 as the primary metric to evaluate the model:
\begin{itemize}[nosep,leftmargin=*]
\item \textbf{Entity}: an entity mention is correct if its offsets and type match a reference entity.
\item \textbf{Relation Strict}: relation with strict match, a relation is correct if its relation type is correct and the offsets and entity types of the related entity mentions are correct.
\item \textbf{Relation Triplet}: relation with boundary match, a relation is correct if its relation type is correct and the string of the subject/object are correct.
\item \textbf{Event Trigger}: an event trigger is correct if its offsets and event type matches a reference trigger.
\item \textbf{Event Argument}: an event argument is correct if its offsets, role type, and event type match a reference argument mention.
\item \textbf{Sentiment Triplet}: a correct triplet requires the offsets boundary of the target, the offsets boundary of the opinion span, and the target sentiment polarity to be all correct at the same time.
\end{itemize}
To make a fair comparison with baseline systems, we mapped the generated string-level extraction results to offset-level for model evaluation.  
In detail, we reconstructed the offset of predicted entity/trigger mentions by finding the matched utterance in the input sequence one by one.
For argument mentions in relation and event tasks, we found the nearest matched utterance to the predicted entity/trigger mention as the predicted offset.
This simple heuristic offset strategy achieves high accuracy.
Compared to the string level evaluation, the error rate of the reported offset level evaluation is less than 0.5\%.
More complicated mapping approaches are left as future work.

\tablename~\ref{tab:hyper-fine-tuning} shows the detailed hyper-parameters for downstream tasks.

\begin{table}[t]
  \centering
    \resizebox{0.49\textwidth}{!}{

    \begin{tabular}{lccccc}
        \toprule
        Methods & PLM   & \textbf{14res} & \textbf{14lap} & \textbf{15res} & \textbf{16res} \\
        \midrule
        \citep{xu-etal-2020-position} & BERT-base & 62.40  & 51.04  & 57.53  & 63.83  \\
        \citep{yan-etal-2021-unified} & BART-base & 65.25  & 58.69  & 59.26  & 67.62  \\
        \citep{xu-etal-2021-learning} & BERT-base & 71.85  & 59.38  & 63.27  & 70.26  \\
        \citep{zhang-etal-2021-towards-generative} & T5-base & 72.16  & 60.78  & 62.10  & 70.10  \\
        \midrule
        \multirow{2}[2]{*}{SSI + SEL} &  UIE-base & \textbf{72.55} & \textbf{62.94} & \textbf{64.41} & \textbf{72.86} \\
              & T5-v1.1-base & 71.27  & 58.69  & 59.60  & 70.24  \\
        \bottomrule
    \end{tabular}%

}
  \caption{Experiment results of UIE-base on the sentiment triplet extraction tasks.}
  \label{tab:absa_base}%
\end{table}%

\begin{table}[t]
  \centering
    \resizebox{0.49\textwidth}{!}{

    \begin{tabular}{lcccc}
    \toprule
    Methods & PLM & \textbf{P} & \textbf{R} & \textbf{F} \\
    \midrule
    \citep{wang-etal-2020-tplinker}	& BERT-base & 91.40 & 92.60 & 92.00 \\
    \citep{sui:2020:spn} & BERT-base & 92.50  & 92.20  & 92.30  \\
    \citep{zheng-etal-2021-prgc} & BERT-base & \textbf{93.50} & 91.90 & \textbf{92.70} \\
    \midrule
    SSI + SEL & T5-v1.1-base &  91.94  & \textbf{93.28} & 92.60  \\
    \bottomrule
    \end{tabular}%

  }
  \caption{Experiment results of SSI and SEL on the NYT (the joint entity and relation extraction setting).}
  \label{tab:nyt_base}
\end{table}

\subsection{Comparison of UIE-base}

This section introduces detailed experiment results of UIE-base.

\tablename~\ref{tab:absa_base} shows the performance of UIE-base and the state-of-the-art systems on the four aspect-based sentiment analysis datasets.
As shown in \tablename~\ref{tab:absa_base}, the proposed SEL and SSI also have strong portability to sentiment triplets extraction, which achieves the competitive performance with the state-of-the-art with task-specific architectures.
With the unified pre-training, UIE-base achieves an improvement of 3.24 on average over T5-v1.1-base across four datasets.
This verifies the proposed unified pre-training algorithms can learn general IE ability even the sentiment knowledge is rarely in the pre-training stage.

\tablename~\ref{tab:nyt_base} shows the performance of SEL-SSI with the T5-v1.1-base for NYT.
Due to the high overlapping of NYT and pre-trained data, we didn't conduct the experiment of UIE on NYT.
Even without pre-training, SSI + SEL still achieved the state-of-the-art performance on NYT.
This is because of the flexible generation architecture and the universal SEL expression, UIE can naturally handle entity overlap problems.

\begin{table*}[htbp]
  \centering
\resizebox{0.99\textwidth}{!}{
\begin{tabular}{ccp{12cm}}
\toprule
\textbf{Task}  & \textbf{Dataset} & \textbf{Structural Schema Instructor} \\
\midrule
Entity & ACE04/05-Ent & <spot> facility <spot> geographical social political <spot> location <spot> organization <spot> person <spot> vehicle <spot> weapon \\
\midrule
Entity & CoNLL03 & <spot> location <spot> miscellaneous <spot> organization <spot> person \\
\midrule
Relation & ACE05-Rel & <spot> facility <spot> geographical social political <spot> location <spot> organization <spot> person <spot> vehicle <spot> weapon <asoc> agent artifact <asoc> general affiliation <asoc> organization affiliation <asoc> part whole <asoc> personal social <asoc> physical \\
\midrule
Relation & CoNLL04 & <spot> location <spot> organization <spot> other <spot> people <asoc> kill <asoc> live in <asoc> located in <asoc> organization in <asoc> work for \\
\midrule
Relation & NYT   & <spot> location <spot> organization <spot> person <asoc> administrative divisions <asoc> advisors <asoc> capital <asoc> children <asoc> company <asoc> contains <asoc> country <asoc> ethnicity <asoc> founders <asoc> geographic distribution <asoc> industry <asoc> location <asoc> major shareholder of <asoc> major shareholders <asoc> nationality <asoc> neighborhood of <asoc> people <asoc> place founded <asoc> place lived <asoc> place of birth <asoc> place of death <asoc> profession <asoc> religion <asoc> teams \\
\midrule
Relation & SciERC & <spot> generic <spot> material <spot> method <spot> metric <spot> other scientific term <spot> task <asoc> compare <asoc> conjunction <asoc> evaluate for <asoc> feature of <asoc> hyponym of <asoc> part of <asoc> used for \\
\midrule
Event & ACE05-Evt & <spot> acquit <spot> appeal <spot> arrest jail <spot> attack <spot> born <spot> charge indict <spot> convict <spot> declare bankruptcy <spot> demonstrate <spot> die <spot> divorce <spot> elect <spot> end organization <spot> end position <spot> execute <spot> extradite <spot> fine <spot> injure <spot> marry <spot> meet <spot> merge organization <spot> nominate <spot> pardon <spot> phone write <spot> release parole <spot> sentence <spot> start organization <spot> start position <spot> sue <spot> transfer money <spot> transfer ownership <spot> transport <spot> trial hearing <asoc> adjudicator <asoc> agent <asoc> artifact <asoc> attacker <asoc> beneficiary <asoc> buyer <asoc> defendant <asoc> destination <asoc> entity <asoc> giver <asoc> instrument <asoc> organization <asoc> origin <asoc> person <asoc> place <asoc> plaintiff <asoc> prosecutor <asoc> recipient <asoc> seller <asoc> target <asoc> vehicle <asoc> victim \\
\midrule
Event & CASIE & <spot> capabilities <spot> common vulnerabilities and exposures <spot> data <spot> databreach <spot> device <spot> discover vulnerability <spot> file <spot> geopolitical entity <spot> malware <spot> money <spot> number <spot> organization <spot> patch <spot> patch vulnerability <spot> payment method <spot> person <spot> personally identifiable information <spot> phishing <spot> purpose <spot> ransom <spot> software <spot> system <spot> time <spot> version <spot> vulnerability <spot> website <asoc> attack pattern <asoc> attacker <asoc> capabilities <asoc> common vulnerabilities and exposures <asoc> compromised data <asoc> damage amount <asoc> discoverer <asoc> issues addressed <asoc> number of data <asoc> number of victim <asoc> patch <asoc> patch number <asoc> payment method <asoc> place <asoc> price <asoc> purpose <asoc> releaser <asoc> supported platform <asoc> time <asoc> tool <asoc> trusted entity <asoc> victim <asoc> vulnerability <asoc> vulnerable system <asoc> vulnerable system owner <asoc> vulnerable system version \\
\midrule
Sentiment & 14/15/16-res & <spot> aspect <spot> opinion <asoc> negative <asoc> neutral <asoc> positive \\
\midrule
Sentiment & 14-lap & <spot> aspect <spot> opinion <asoc> negative <asoc> neutral <asoc> positive \\
\bottomrule
\end{tabular}%

}
  \caption{Structured schema instructor for each dataset (we use <spot> and <asoc> rather than [spot] and [asoc] for better visualization).}
  \label{tab:addlabel}%
\end{table*}%

\begin{table*}[htbp]
  \centering

\resizebox{0.99\textwidth}{!}{
  \begin{tabular}{ccl}
  \toprule
  \textbf{Task} & \textbf{Dataset} & \textbf{Structured Extraction Language} \\
  \midrule
  Entity & ACE04/ACE05-Ent & \texttt{((geographical social political:~Filipino)} \\
        &       & \texttt{~(person:~Filipino President)} \\
        &       & \texttt{~(person:~Filipino President Ramos)} \\
        &       & \texttt{~(person:~the six people awarded Magasaysay award)} \\
        &       & \texttt{~(person:~Magasaysay))} \\
  \midrule
  Entity & CoNLL03 & \texttt{((organization:~EU)} \\
        &       & \texttt{~(miscellaneous:~German)} \\
        &       & \texttt{~(miscellaneous:~British))} \\
  \midrule
  Relation & ACE05-Rel & \texttt{((geographical social political:~European)} \\
        &       & \texttt{~(geographical social political:~troika} \\
        &       & \texttt{~~(part whole:~European))} \\
        &       & \texttt{~(geographical social political:~itself)} \\
        &       & \texttt{~(geographical social political:~Washington))} \\
  \midrule
  Relation & CoNLL04 & \texttt{((location:~Rome} \\
        &       & \texttt{~~(located in:~Lazio))} \\
        &       & \texttt{~(location:~Lazio)} \\
        &       & \texttt{~(location:~Naples} \\
        &       & \texttt{~~(located in:~Campania))} \\
        &       & \texttt{~(location:~Campania))} \\
  \midrule
  Relation & NYT   & \texttt{((person:~William F. Weld} \\
        &       & \texttt{~~(place lived:~New York))} \\
        &       & \texttt{~(location:~New York))} \\
  \midrule
  Relation & SciERC & \texttt{((method:~HMMs)} \\
        &       & \texttt{~(other scientific term:~weak duration constraints} \\
        &       & \texttt{~~(feature of:~HMMs)))} \\
  \midrule
  Event & ACE05-Evt & \texttt{((transport:~heading} \\
        &       & \texttt{~~(artifact:~family)} \\
        &       & \texttt{~~(destination:~new hampshire)} \\
        &       & \texttt{~~(origin:~lakeland)} \\
        &       & \texttt{~~(vehicle:~plane)))} \\
  \midrule
  Event & CASIE & \texttt{((phishing:~email scam} \\
        &       & \texttt{~~(trusted entity:~a Netflix notification)} \\
        &       & \texttt{~~(victim:~subscribers)} \\
        &       & \texttt{~~(trusted entity:~the streaming service))} \\
        &       & \texttt{~(file:~a Netflix notification)} \\
        &       & \texttt{~(person:~subscribers)} \\
        &       & \texttt{~(system:~the streaming service))} \\
  \midrule
  Sentiment & 14/15/16-res & \texttt{((aspect:~staff} \\
        &       & \texttt{~~(negative:~horrible))} \\
        &       & \texttt{~(opinion:~horrible))} \\
  \midrule
  Sentiment & 14lap & \texttt{((opinion:~good)} \\
        &       & \texttt{~(aspect:~battery life} \\
        &       & \texttt{~~(positive:~good)))} \\
  \bottomrule
  \end{tabular}%
}
  \caption{Structured extraction language expressions for each dataset.}
  \label{tab:sel-example}%
\end{table*}%

\end{document}